\documentclass[letterpaper]{article} % DO NOT CHANGE THIS
\usepackage{aaai2026}  % DO NOT CHANGE THIS
\usepackage{times}  % DO NOT CHANGE THIS
\usepackage{helvet}  % DO NOT CHANGE THIS
\usepackage{courier}  % DO NOT CHANGE THIS
\usepackage[hyphens]{url}  % DO NOT CHANGE THIS
\usepackage{graphicx} % DO NOT CHANGE THIS
\usepackage{natbib}  % DO NOT CHANGE THIS AND DO NOT ADD ANY OPTIONS TO IT
\usepackage{caption} % DO NOT CHANGE THIS AND DO NOT ADD ANY OPTIONS TO IT
\usepackage{algorithm}
\usepackage{algpseudocode}
\usepackage{amsmath,amsfonts}
\usepackage{newfloat}
\usepackage{listings}
\DeclareCaptionStyle{ruled}{labelfont=normalfont,labelsep=colon,strut=off} % DO NOT CHANGE THIS
\floatstyle{ruled}
\newfloat{listing}{tb}{lst}{}
\floatname{listing}{Listing}
\title{KineST: A Kinematics-guided Spatiotemporal State Space Model for Human Motion Tracking from Sparse Signals}
\author{
    Shuting Zhao\textsuperscript{\rm 1, 3}\equalcontrib, Zeyu Xiao\textsuperscript{\rm 2}\equalcontrib, Xinrong Chen\textsuperscript{\rm 1, 3}\thanks{Corresponding author}
}
\affiliations{
    \textsuperscript{\rm 1}College of Biomedical Engineering, Fudan University\\

    \textsuperscript{\rm 2}College of Intelligent Robotics and Advanced Manufacturing, Fudan University\\
    \textsuperscript{\rm 3}Shanghai Key Laboratory of Medical Image Computing and Computer
Assisted Intervention \\
    zhaoshuting@fudan.edu.cn, xiaozy23@m.fudan.edu.cn, chenxinrong@fudan.edu.cn
}

\usepackage{bibentry}
\begin{document}

\maketitle

\begin{abstract}
Full-body motion tracking plays an essential role in AR/VR applications, bridging physical and virtual interactions. However, it is challenging to reconstruct realistic and diverse full-body poses based on sparse signals obtained by head-mounted displays, which are the main devices in AR/VR scenarios. Existing methods for pose reconstruction often incur high computational costs or rely on separately modeling spatial and temporal dependencies, making it difficult to balance accuracy, temporal coherence, and efficiency. To address this problem, we propose KineST, a novel kinematics-guided state space model, which effectively extracts spatiotemporal dependencies while integrating local and global pose perception. The innovation comes from two core ideas. Firstly, in order to better capture intricate joint relationships, the scanning strategy within the State Space Duality framework is reformulated into kinematics-guided bidirectional scanning, which embeds kinematic priors. Secondly, a mixed spatiotemporal representation learning approach is employed to tightly couple spatial and temporal contexts, balancing accuracy and smoothness. Additionally, a geometric angular velocity loss is introduced to impose physically meaningful constraints on rotational variations for further improving motion stability. Extensive experiments demonstrate that KineST has superior performance in both accuracy and temporal consistency within a lightweight framework. Project page: \url{https://kaka-1314.github.io/KineST/}

\end{abstract}

% Links section - only shown in camera-ready version
% \ifdefined\aaaianonymous

% Version-specific content
% \ifdefined\aaaianonymous
\section{Introduction}
% 任务很重要，但实现困难
Full-body pose reconstruction based on Head-Mounted Displays (HMDs) facilitates a diverse array of AR/VR applications, including patient rehabilitation, realistic avatar generation, and the control of teleoperated humanoid robots~\cite{1,66}. However, due to the sparsity of signals captured by HMDs, inferring accurate and natural full-body motion remains a challenging problem. 
%[1] Full-body pose reconstruction and correction in virtual reality for rehabilitation training [2] robots

\begin{figure}
    \centering
    \includegraphics[width=1\linewidth]{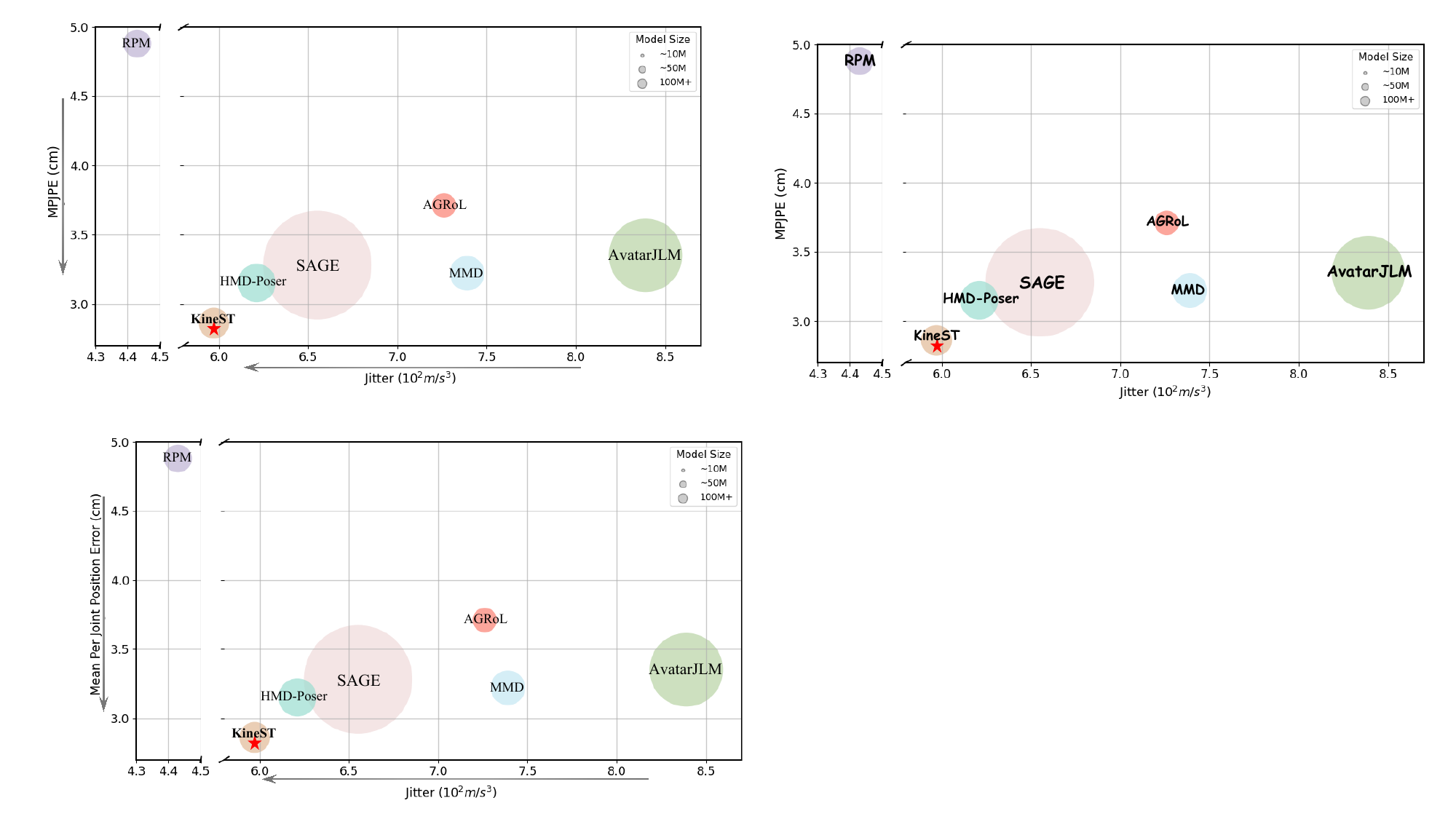}
        \caption{Comparison of our approach with state-of-the-art methods in terms of overall performance. Our method achieves the smallest average position error and smoother motion, and maintains a lightweight model architecture.}
    \label{fig:Teaser}
\end{figure}

% 之前一些工作能够较好得重建人体姿态，但却付出了高昂的计算和存储的代价。（对于VR场景，便携高效很重要）
Previous works have demonstrated the feasibility of reconstructing realistic full-body motion, but often at the expense of substantial computational cost and large parameter counts, which limits its application. For example, AvatarJLM~\cite{5} improves performance by stacking multiple Transformer blocks~\cite{19} in a high-dimensional space, while SAGE~\cite{7} leverages large generative models such as VQ-VAE~\cite{16} and diffusion models~\cite{67}. Their high deployment costs underline the need for more efficient solutions that can achieve high accuracy with a compact framework. 

%[3] AvatarJlM [5] SAGE

% 为了设计出更加轻巧的模型，最近一些工作做出了努力 （因为输入稀疏，从轻量模型中拟合逼真的全身人体姿态，很难）

To build lightweight and robust models for realistic full-body pose estimation from sparse inputs, recent works have explored various solutions. For example, RPM~\cite{70} introduces a prediction consistency anchor to reduce sudden motion drift, which improves smoothness but sacrifices pose accuracy. To improve pose accuracy, separate temporal and spatial modules are adopted to better capture the complex dependencies of human motion~\cite{69,71}. Although this dual-module design enhances joint interaction modeling, some modeling capacities are shifted to single-frame spatial features, which can compromise motion smoothness. Therefore, a key question is raised: \textbf{how can we design a model that remains lightweight yet achieves both high accuracy and motion smoothness?} 

%~\cite{2,68,69,70,71}
% 最近mamba的提出，给该任务带来了新的机遇，但是直接用效果不好。
Recently, the State Space Duality (SSD) framework~\cite{8} introduces a special and robust scanning strategy and shows great promise for efficient time-series modeling, making it a strong candidate for our task. However, directly applying SSD to human motion tracking yields unsatisfactory performance, primarily due to its unidirectional scanning and the absence of specific designs for full-body pose reconstruction.

% To tackle this challenge, we propose xxx
To tackle this challenge, we propose KineST, a lightweight yet effective model, to fully extract spatiotemporal dependencies while integrating local and global pose information. Specifically, we first design a Temporal Flow Module (TFM) to learn inter-frame dynamics. The main components in TFM are SSD blocks in which a bidirectional scanning strategy is employed to initially capture motion features. These are followed by a Local Motion Aggregator (LMA) and a Global Motion Aggregator (GMA), which progressively refine local and global motion dependencies. 

Secondly, we introduce a robust Spatiotemporal Kinematic Flow Module (SKFM), which employs a Spatiotemporal Mixing Mechanism (STMM) to tightly couple spatial and temporal contexts and maintain a balance between accuracy and temporal continuity. Moreover, a novel Kinematic Tree Scanning Strategy (KTSS) is employed to incorporate kinematic priors into spatial feature capture and fully capture intricate joint relations. To further improve motion continuity, a geometric angular velocity loss is proposed, jointly constraining both the magnitude and direction of rotation changes in a geometrically consistent way. The contributions of this work are summarized as follows:

\begin{itemize}
    \item A kinematics-guided state space model, KineST, is proposed, which can not only fully extract spatiotemporal information but also integrate local-global perception.
    \item A robust Spatiotemporal Kinematic Flow Module (SKFM) is designed, which applies the Spatiotemporal Mixing Mechanism (STMM) to tightly couple spatial and temporal contexts and employs a novel Kinematic Tree Scanning Strategy (KTSS) to fully capture intricate joint relations.
    \item To further improve motion continuity, a geometric angular velocity loss is proposed to impose physically meaningful constraints on rotational variations.
    \item Extensive experiments demonstrate KineST’s superiority over current state-of-the-art methods. Furthermore, comprehensive ablation studies confirm the contribution of each carefully designed component.
\end{itemize}

\section{Related Works}
\subsection{Motion Tracking from Sparse Observations }
Earlier methods explore full-body tracking using 4 or 6 IMUs~\cite{4,12,13,14,140,15}. However, in AR/VR scenarios, HMDs are more practical and widely adopted, which typically provide only 3 tracking signals from the head and hands. Based on HMD inputs, generative techniques are adopted to synthesize full-body poses. For example, in FLAG~\cite{18} and VAE-HMD~\cite{17}, a variational auto-encoder (VAE) and a flow-based model are applied, respectively. In AGRoL~\cite{6} and SAGE~\cite{7}, the reconstruction of avatars is achieved by diffusion models or VQ-VAEs. Correspondingly, another type of work is based on regression-based approaches. Transformer-based architecture is adopted to predict full-body poses from these three sparse signals, such as AvatarPoser~\cite{3}, AvatarJLM~\cite{5}, HMD-Poser~\cite{69}, and RPM~\cite{70}. KCTD~\cite{2} designs an MLP-based model with kinematic constraints for the task. With the recent advancements in state space models (SSMs), several studies have explored their potential for this task. For instance, MMD~\cite{71} leverages the sequential modeling capability of SSMs to track full-body poses in the temporal and spatial dimensions, respectively. SSD-Poser~\cite{68} further introduces a hybrid architecture combining SSD and Transformers to efficiently capture contextual motion features. 

Although promising results are achieved, they struggle to balance pose accuracy and motion smoothness within a lightweight framework. To address this, a kinematics-guided spatiotemporal state space model, KineST, is designed to improve both accuracy and continuity.

\subsection{Spatial Modeling in Human Pose Estimation}

Spatial modeling plays a vital role in full-body motion reconstruction from sparse signals, where accurately capturing inter-joint relationships within a single frame is crucial for reconstructing plausible poses. While spatial modeling has been widely explored in related domains such as action recognition and image-based pose estimation~\cite{76,77,78,79,80,81,82}, its application under sparse input constraints presents unique challenges. In most existing methods~\cite{5,72,69,75}, self-attention mechanisms are employed to treat the 22 body joints as independent tokens and learn pairwise dependencies through similarity weights(Figure~\ref{scan}(b)). In contrast, BPG~\cite{74} employs a GCN-based structure to explicitly encode local joint connectivity via an adjacency matrix, while MMD~\cite{71} leverages SSMs to process concatenated joint features as a whole, implicitly modeling global joint correlations. 
%,83

Despite their effectiveness, these approaches focus on either local or global relations, and often fail to tightly integrate spatial and temporal contexts. To address this limitation, we introduce KTSS to enhance the spatial modeling capacity based on body topology, and further propose STMM to jointly encode spatial-temporal dependencies, striking a better balance between accuracy and smoothness.

%DiffPose: SpatioTemporal Diffusion Model for Video-Based HumanPose Estimation
%SpatioTemporal Learning for Human Pose Estimation in Sparsely-Labeled Videos
%Spatio-Temporal Fusion for Human Action Recognition via Joint Trajectory Graph

\begin{figure*}[t]
\centering
\includegraphics[width=1\textwidth]{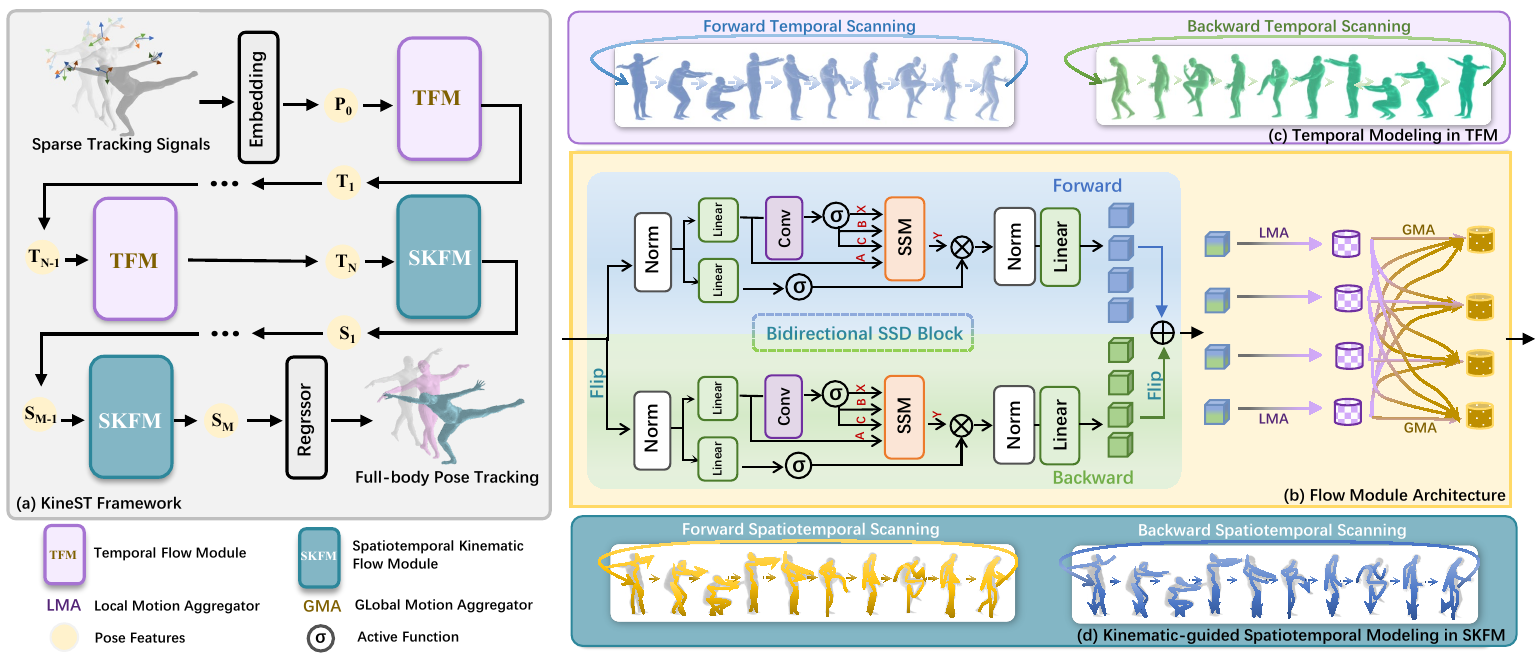} % Reduce the figure size so that it is slightly narrower than the column.
\caption{Overall architecture. (a) The architecture of the proposed KineST model, whose main components are the temporal flow module (TFM) and spatiotemporal kinematic flow module (SKFM). (b) The shared structure of the flow module used in both TFM and SKFM, which comprises a bidirectional SSD block, a local motion aggregator (LMA), and a global motion aggregator (GMA). (c) Temporal modeling within the TFM. (d) Kinematics-guided spatiotemporal modeling within the SKFM.  }
\label{fig1}
\end{figure*}

% \ifdefined\aaaianonymous

\section{Preliminary}

\subsubsection{State Space Duality} The SSD framework~\cite{8} is an advanced and efficient variant of SSMs, offering enhanced capabilities in both computation and inference. This framework introduces a novel matrix-based computation structure, which integrates the linear recurrence properties of SSMs with a quadratic dual formulation. Specifically, the SSD framework is denoted as follows:
\begin{equation}
\begin{gathered}
\label{eq.5}
y_t=\sum_{i=0}^t C_t^T A_{t: i}^X B_i x_i, \\
y=S S M(A, B, C)(x)=M x,
\end{gathered}
\end{equation}
where $A_{t: i}^X $ denotes the product of ${A}$ terms from $i$ + 1 to $t$, and $M$ is defined as:$M_{j i}:=C_j^{\top} A_j \cdots A_{i+1} B_i$. When $A_i $ is reduced to a scalar, Eq.\eqref{eq.5} can be reformulated as:
\begin{equation}
\begin{gathered}
{y}={Mx}={F} \cdot\left({C}^{{T}} {~B}\right) {x}, \\
where {F}_{{ji}}= \begin{cases}A_{j} A_{j-1}  \cdots  A_{i+1} & i<j \\ 1 & i=j \\ 0 & i>j\end{cases}.
\end{gathered}
\end{equation}In SSD framework, the original time-series recurrence is reformulated into an equivalent product-sum matrix form, which can be efficiently parallelized by associative scan algorithms~\cite{84,85}. In this work, we extend this scan mechanism by embedding kinematic priors and bidirectional constraints to enhance realistic full-body pose reconstruction. 
%[93] Jimmy TH Smith, Andrew Warrington, and Scott W Linderman. “Simplified State Space Layers for Sequence Modeling”. In: The International Conference on Learning Representations (ICLR). 2023.
%Jimmy TH Smith, Andrew Warrington, and Scott W Linderman. “Simplified State Space Layers for Sequence Modeling”. In: The International Conference on Learning Representations (ICLR). 2023.

\section{Method}
\subsection{Problem Formulation}
The purpose of this task is to achieve realistic full-body motion prediction 
$Y=\left\{ y_i \right\}_{i=1}^L \in \mathbb{R}^{L \times V}$ 
from sparse IMU signals 
$X=\left\{ x_i \right\}_{i=1}^L \in \mathbb{R}^{L \times C}$ 
captured from the head and hands over $L$ time frames, 
where $C$ and $V$ denote the input and output dimensions, respectively. 
The ground-truth full-body motion is denoted as 
$Z = \left\{ z_i \right\}_{i=1}^L \in \mathbb{R}^{L \times V}$. Following~\cite{3}, each $\mathbf{x}_i$ contains a 3D position, 6D rotation, linear velocity, and angular velocity for each of the three tracked parts. We adopt pose parameters of the first 22 joints of the SMPL model~\cite{29} to represent the output. As a result, the input and output dimensions are $C=3 \times (3+6+3+6)$ and $V = 22 \times 6 $.

\subsection{Overall architecture}

The overall architecture of the proposed KineST model consists of Temporal Flow Modules (TFMs) and Spatiotemporal Kinematic Flow Modules (SKFMs), as shown in Figure~\ref{fig1} (a). Given the sparse tracking signals ${X} \in \mathbb{R}^{L \times C}$, we first use a single linear layer to obtain embedded pose features ${P_0} \in \mathbb{R}^{L \times E}$. Here $C$ and $E$ represent the feature dimensions of the original and embedded signals, respectively. Subsequently, the ${P_0} \in \mathbb{R}^{L \times E}$ are processed through a stack of $N$ TFMs, producing deep temporal motion features ${T_n} \in \mathbb{R}^{L \times E}$ at each stage, where $n\in\left\{{1, 2, ..., N}\right\}$. To further learn kinematics-guided spatiotemporal information, the output ${T_N} \in \mathbb{R}^{L \times E}$ is passed through a sequence of $M$ SKFM blocks. The ${S}_m$ is produced at each stage, where $m\in\left\{{1, 2, ..., M}\right\}$. Finally, the full-body motion poses are estimated by a linear regressor.

% the output ${T_N} \in \mathbb{R}^{L \times E}$ is transformed into the mixed tensor ${S_0'} \in \mathbb{R}^{L \times J \times D}$, where $J$ denotes the number of full-body joints and $D$ denotes the latent feature dimension for each joint.

\begin{figure}[t]
    \centering
    \includegraphics[width=1\linewidth]{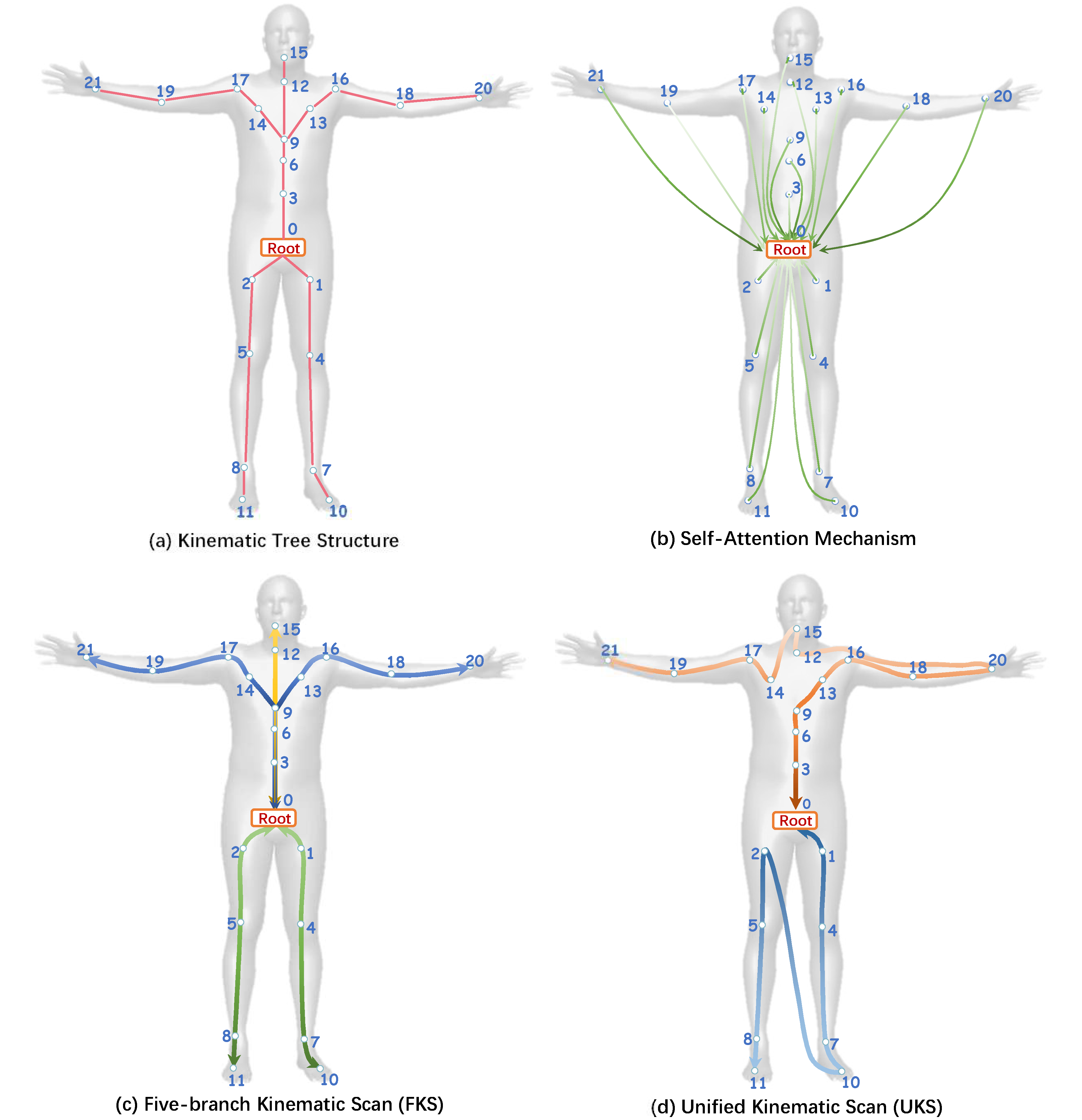}
    \caption{Comparison of different scanning strategies. }
    \label{scan}
\end{figure}

\subsection{Temporal Flow Module}
To learn inter-frame motion dynamics, we propose the Temporal Flow Module (TFM) whose architecture is shown in Figure~\ref{fig1}(b). It adopts a bidirectional SSD block (Bi-SSD) with parallel forward and backward branches to enhance temporal modeling. Each SSD block comprises layer normalization (LN), linear layers, convolutional layers, the SiLU activation function, and the core state space model (SSM). Given embedded features ${P_0} \in \mathbb{R}^{L \times E}$, the forward branch is formulated as:
\begin{equation}
\begin{gathered}
\mathrm{X, B, C} = \mathrm{SiLU}(\mathrm{Conv}(\mathrm{Linear}(\mathrm{LN}(\mathrm{P_0})))),\\
\mathrm{A} = \mathrm{Linear}(\mathrm{LN}(\mathrm{P_0})),\\
f_1 = \mathrm{SiLU}(\mathrm{Linear}(\mathrm{LN}(\mathrm{P_0}))),\\
\mathrm{F}_f^t = \mathrm{Linear}(\mathrm{LN}(f_1 \odot \mathrm{SSM}(X, A, B, C))),
\end{gathered}
\end{equation}
where $\mathrm{X}$, $\mathrm{A}$, $\mathrm{B}$ and $\mathrm{C}$ denote the state vector, state transition matrix, input matrix, and output matrix, respectively, while $f_1$ serves as an adaptive gating vector and $F_f^t$ represents forward temporal motion features. The backward branch shares the same structure, except that ${P_0}$ is time-reversed to obtain ${P_0^{flip}}$, passed through the same operations, and flipped back to yield backward features $F_b^t$. To enhance representation, we further apply a Local Motion Aggregator (LMA) and a Global Motion Aggregator (GMA) for modeling local dependencies and global motion periodicity, respectively:
\begin{equation}
\mathrm{T}_1 = \mathrm{GMA}(\mathrm{LMA}(F_f^t + F_b^t)).
\end{equation}
LMA is implemented via convolution, while GMA uses a lightweight transformer. More details are provided in the supplementary material.

\begin{table*}[t]
    \centering
    \caption{Evaluation results on three subsets of AMASS under Protocol 1. The best results are in bold, and the second-best results are underlined. For fair comparison, HMD-Poser and RPM are retrained using their public code, as they originally use different body shape parameters or FPS, and are denoted with *.}
    \label{tab:Table1}
    % \scriptsize
    \small
    \setlength{\tabcolsep}{4pt}
    \begin{tabular}{l|ccccccccc}
\hline Method & MPJRE↓ & MPJPE↓ & MPJVE↓ & Hand PE↓ & Upper PE↓ & Lower PE↓ & Root PE↓ & Jitter↓ & Param. \\
\hline  AvatarPoser~\cite{3} & 3.08 & 4.18 & 27.70 & 2.12 & 1.81 & 7.59 & 3.34 & 14.49 & 4M \\ 
        AvatarJLM~\cite{5} & 2.90 & 3.35 & 20.79 & 1.24 & 1.72 & 6.20 & 2.94 & 8.39 & 63M \\
        AGRoL~\cite{6} & 2.66 & 3.71 & 18.59 & 1.31 & 1.55 & 6.84 & 3.36 & {7.26} & 7M\\
        SAGE~\cite{7} & 2.53 & 3.28 & 20.62 & 1.18 & 1.39 & 6.01 & 2.95 & 6.55 & 137M \\
        HMD-Poser*~\cite{69} & 2.32 & \underline{3.15} & 18.15 & 1.35 & 1.34 & \underline{5.76} & \underline{2.76} & 6.21 & 17M\\
        MMD~\cite{71} & \underline{2.31} & 3.22 & \underline{17.88} & \textbf{0.94} & \underline{1.29} & 6.01 & 2.88 & 7.39 & 14M \\
        RPM*~\cite{70} & 3.69 & 4.88 & 21.99 & 5.90 & 2.94 & 7.69 & 4.01 & \textbf{4.43} & 9M \\
        % \rowcolor[gray]{0.94} % 设置浅灰色背景
        \textbf{KineST (Ours)} & \textbf{2.25} & \textbf{2.86} & \textbf{15.26} & \underline{1.04}& \textbf{1.24} & \textbf{5.20} & \textbf{2.65} & \underline{5.97}& 11M \\
        \hline 
        GT & 0 & 0 & 0 & 0 & 0 & 0 & 0 & 4.00 & ——\\
\hline
    \end{tabular}
\end{table*}

\begin{figure*}[t]
    \centering
    \includegraphics[width=1\linewidth]{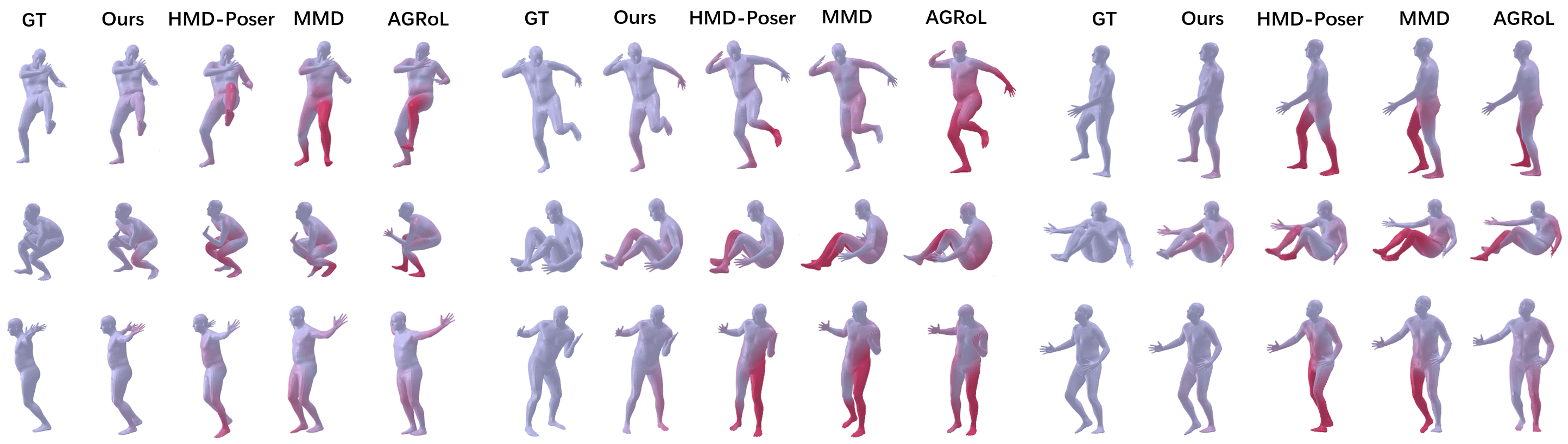}
    \caption{Visualization results of different actions compared with other methods. The joint error degrees are indicated by red shading, allowing a comparative assessment of reconstruction accuracy across various poses for each method. These visuals confirm the robustness and enhancements of the proposed model, particularly in lower body predictions.}
    \label{fig.single_pose}
\end{figure*}

\subsection{Spatiotemporal Kinematic Flow Module }

To better extract inter-joint dependencies while preserving motion smoothness, we propose the Spatiotemporal Kinematic Flow Module (SKFM), which performs kinematics-guided spatiotemporal modeling. Specifically, a Kinematic Tree Scanning Strategy (KTSS) is introduced to inject kinematic priors into spatial feature extraction enhancing joint interactions, while a Spatiotemporal Mixing Mechanism (STMM) is incorporated to tightly couple spatial and temporal features to balance pose accuracy and continuity. It is worth noting that SKFM shares the same overall structure as TFM, as shown in Figure~\ref{fig1}(b).

% \begin{figure}[t]
%     \centering
%     \includegraphics[width=1\linewidth]{scan2.png}
%     \caption{Comparison of different scanning strategies. }
%     \label{scan}
% \end{figure}
%(a) The kinematic tree structure defined in SMPL. (b) Self-attention methods. (c) The proposed Five-branch Kinematic Scan (FKS). (d) The proposed Unified Kinematic Scan (UKS).

\begin{algorithm}[tb]
\caption{Spatiotemporal Mixing Mechanism}
\label{alg:algorithm}
\textbf{Input}: Final temporal motion features $\boldsymbol{T}_N \in \mathbb{R}^{L \times E}$\\
\textbf{Output}: Final spatiotemporal motion features $\boldsymbol{S}_M \in \mathbb{R}^{L \times E}$ \\
\begin{algorithmic}[1]
\For{i in $M$}
    \If{$i = 0$} $\boldsymbol{S}_l \leftarrow \text{Linear}(\boldsymbol{T}_N)$
\Else $\ \boldsymbol{S}_l \leftarrow \text{Linear}(\boldsymbol{S}_i)$
\EndIf
    \State $\boldsymbol{S}_l' \leftarrow \text{rearrange}(\boldsymbol{S}_l,\ L\ F \rightarrow L\ (J\ D))$
    \State $\boldsymbol{S}_f,\ \boldsymbol{S}_b \leftarrow \text{KTSS}(\boldsymbol{S}_l')$
    \State $\boldsymbol{S}'_f \leftarrow \text{rearrange}\left( \boldsymbol{S}_f ,\ L\ J_f\ D \rightarrow (L\ J_f)\ D \right)$
    \State $\boldsymbol{S}'_b \leftarrow \text{rearrange}\left( \boldsymbol{S}_b ,\ L\ J_b\ D \rightarrow (L\ J_b)\ D \right)$
    \State $\boldsymbol{S}''_f,\ \boldsymbol{S}''_b \leftarrow \text{Bi-SSD}(\boldsymbol{S}'_f,\ \boldsymbol{S}'_b)$
    \State $\boldsymbol{S}_{i+1} \leftarrow \text{GMA}(\text{LMA}(\text{Linear}(\boldsymbol{S}''_f + \boldsymbol{S}''_b)$))
\EndFor
\State \textbf{return} $\boldsymbol{S}_M$
\end{algorithmic}
\end{algorithm}
\subsubsection{Kinematic Tree Scanning Strategy}
In the SMPL model, the human skeleton is represented as a hierarchical structure (Figure~\ref{scan} (a)), where a kinematic tree encodes the parent-child relationship between joints. Such kinematic representations are essential for realistic motion tracking and allow efficient control over joint transformations.

% In the SMPL model~\cite{29}, the human skeleton is organized as a kinematic tree (Fig.~\ref{scan}(a)) that encodes parent-child joint relationships. This hierarchical structure enables realistic motion tracking and efficient joint control.

However, existing methods overlook these important kinematic priors, leading to suboptimal performance. In this paper, we leverage the sequential nature of SSD, which allows each joint feature to be inferred from the previous joint state. By reformulating the original unidirectional scan into a kinematics-guided bidirectional scan, the interactions between the parent-child joints are effectively captured, enabling features to flow forward and backward along the kinematic hierarchy. Within this design, we introduce two distinct scanning variants under this framework for efficient full-body reconstruction. 

First, we introduce the Five-branch Kinematic Scan (FKS), which strictly follows the kinematic tree structure, as shown in Figure~\ref{scan}(c). The forward scanning order is [0,1,4,7,10,0,2,5,8,11,0,3,6,9,13,16,18,20, 0,3,6,9,12,15,0,3,6,9,14,17,19,21]. Compare to the index-order scan in SMPL (sequentially from 0 to 21), FKS allows better perception of local kinematic dependencies. However, its branch-wise design limits the model's ability to capture integral body movement dynamics (Table~\ref{tab:Table4}). 

To address these issues, we propose the Unified Kinematic Scan (UKS), as shown in Figure~\ref{scan}(d). By positioning the root joint centrally, UKS effectively couples upper and lower body motion, enhancing global motion coherence and overall reconstruction robustness. The specific forward scanning order is [21,19,17,14,15,12,20,18,16,13,9,6,3,0,1,4,7,10,2,5,8,11]. Therefore, the proposed KTSS mainly adopts UKS to more effectively capture full-body joint dependencies. 

\begin{figure*}[htb]
    \centering
    \includegraphics[width=1\linewidth]{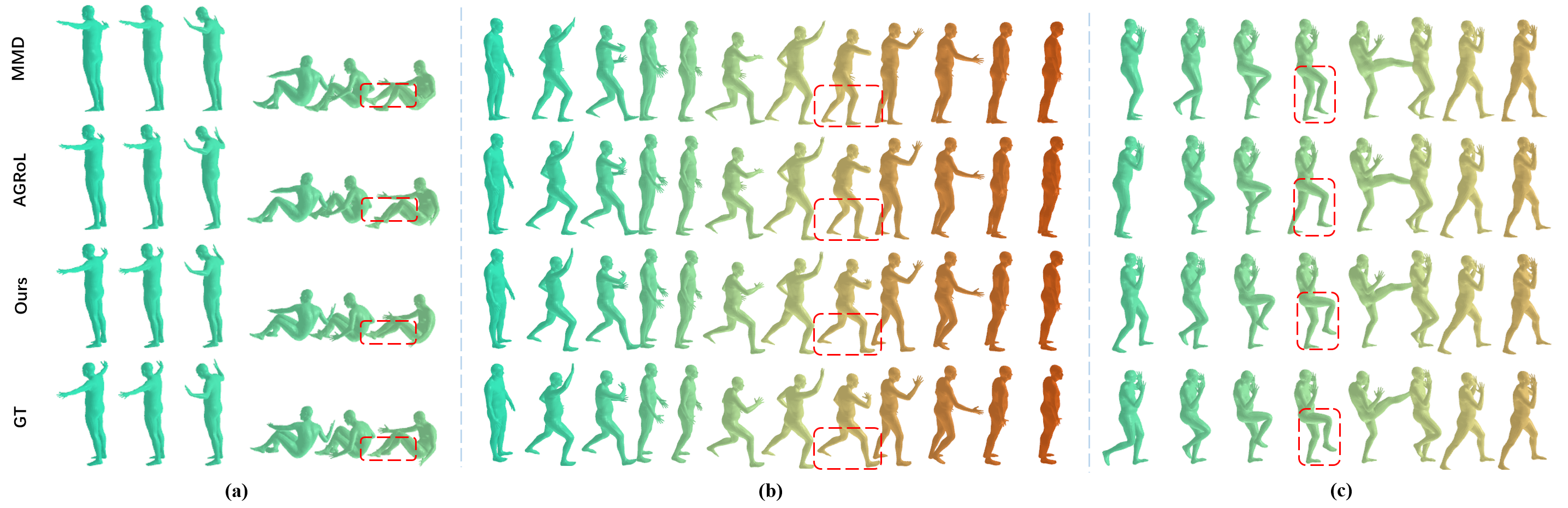}
    \caption{Visualization results of continuous pose sequences compared with other methods. The visualization illustrates that the proposed model delivers smoother and more realistic body motion tracking. Notably, the proposed model provides refined reconstruction highlighted by red dashed boxes.}
    \label{fig:seq_pose}
\end{figure*}

\subsubsection{Spatiotemporal Mixing Mechanism }
%加上伪代码在附录
To capture inter-joint relations and maintain smoothness over frames, the Spatiotemporal Mixing Mechanism (STMM) is employed through mixed spatiotemporal representation learning. The process of STMM is presented in Algorithm~\ref{alg:algorithm}. Specially, we start from the final temporal motion features ${T_N} \in \mathbb{R}^{L \times E}$, which are projected into a latent joint space ${S_l} \in \mathbb{R}^{L \times H}$ and then reshaped into detailed joint features ${S_l'} \in \mathbb{R}^{L \times J \times D}$, where $H = J \times D$. Following KTSS, the joints are reordered along both forward and backward directions to align features according to the kinematic chain, resulting in two new tensors: ${S_f} \in \mathbb{R}^{L \times J_f \times D}$ and ${S_b} \in \mathbb{R}^{L \times J_b \times D}$, where $J_f$ and $J_b$ denote the forward and backward joint sequences, respectively. 

To perceive spatial and temporal features, we rearrange the sequence and joint dimensions into a unified axis, obtaining two mixed tensors: ${S}_f' \in \mathbb{R}^{(L J_f) \times D}$ and ${S}_b' \in \mathbb{R}^{(L  J_b) \times D}$. These tensors are then processed by the Bi-SSD to enhance spatiotemporal dependencies. Finally, the bidirectional features are summed and linearly projected, followed by sequential aggregation via LMA and GMA, to produce the refined spatiotemporal features ${S_1} \in \mathbb{R}^{L \times E}$. After $M$ iterations of processing, the final spatiotemporal motion features ${S_M} \in \mathbb{R}^{L \times E}$ are obtained, preparing for regression.

%However, under sparse input signals, these losses provide limited temporal regularization, failing to adequately constrain the inter-frame continuity. To mitigate this issue, existing works~\cite{70,72} introduce angular velocity regularization, typically estimating angular velocity by applying first-order finite differences to rotation representations. However, since rotations lie on the nonlinear manifold of the Lie group $\mathrm{SO}(3)$, angular velocity should be computed within its tangent space, namely the Lie algebra $\mathfrak{so}(3)$~\cite{73}. Therefore, we propose a geometric angular velocity loss $\mathcal{L}_{\text{angvel}}^{\text{geo}}$ to physically capture the true geometric relationship between rotations. 

\subsection{Loss Function}
We retain two commonly used supervision terms, the L1 loss on body orientation and the L1 loss on joint rotations~\cite{3}. To further enhance motion continuity, a geometric angular velocity loss $\mathcal{L}_{\text{angvel}}^{\text{geo}}$ is proposed to physically capture the true geometric relationship between rotations. Unlike existing works~\cite{70,72} estimating angular velocity by applying first-order finite differences to rotation representations, we compute angular velocity within its tangent space, namely the Lie algebra $\mathfrak{so}(3)$~\cite{73}. Further details are provided in the supplementary materials. 

Specifically, we begin by defining the angular velocity at time $\text{t}$ as the geodesic rotational difference between two consecutive frames, formulated as $ V_t = R_{t-1}^{-1}R_t$. Here, $R_t$ is the rotation matrix derived from $z_t$ via Gram-Schmidt orthogonalization. To accurately measure the difference between two angular velocities, we impose constraints from two perspectives: rotational magnitude and rotational direction. The rotational magnitude is measured using a natural Riemannian metric defined on the compact Lie group $\mathrm{SO}(3)$, while the rotational direction is extracted from the skew-symmetric part of the relative rotation matrix. Each relative rotation is eventually mapped to its corresponding axis-angle representation, and the formulation is defined as follows:
\begin{equation}
\begin{gathered}
\theta_V = \arccos\left( \frac{\mathrm{Tr}(V) - 1}{2} \right),\\
\log V = \theta_V \cdot \frac{1}{2 \sin \theta_V}
\begin{bmatrix}
V_{32} - V_{23} \\
V_{13} - V_{31} \\
V_{21} - V_{12}
\end{bmatrix},\\
\mathcal{L}_{\text{angvel}}^{\text{geo}} = \sum_{t=1}^{T-1} \left\| \log(V_t) - \log(\hat{V}_t) \right\|_1.\\
\end{gathered}
\end{equation}
Here, $V_t$ denotes the GT angular velocity at frame $t$, $\hat{V}_t$ represents the corresponding predicted angular velocity computed from $y_t$, $\theta_V$ denotes the rotational magnitude of $V$. 

In summary, the final loss function is defined as:
$$
\mathcal{L} = \alpha \mathcal{L}_{\text{rot}} + \beta \mathcal{L}_{\text{ori}} + \delta \mathcal{L}_{\text{angvel}}^{\text{geo}}.
$$
The weights for each component are empirically set to $\alpha = 1$, $\beta = 0.02$, and $\delta = 1$, respectively.

% \begin{figure*}
%     \centering
%         \centering        \includegraphics[width=1\linewidth]{seq_pose2.png}
%         \caption{Enter Caption}
%         \label{fig:enter-label}
% \end{figure*}

\section{Experiments}
% a large human motion database, which unifies different motion capture datasets~\cite{33,34,35,40,41,42,43,44,45,47,48,49,47}
\subsection{Dataset and Implementation Details}

Our method is trained and evaluated on the AMASS~\cite{10} and a real-captured dataset~\cite{5}, both represented using SMPL parameters. The TFMs and SKFMs each contain 2 modules, with an embedded feature dimension $E = 256$. The number of full-body joints $J$ is 22, each with a latent dimension $D = 64$, and the input sequence length $L = 96$. Training is conducted on an NVIDIA 4090 with a batch size of 256. We adopt the Adam optimizer~\cite{37} with a learning rate of 3e-4 (decayed to 3e-5 after 200000 iterations) and a weight decay of 1e-5. Inference is efficient, requiring only 12.9 ms to process 96 frames on the same GPU.

% The proposed method is trained and tested on the AMASS~\cite{10} dataset and real-captured dataset~\cite{5}, which are represented by SMPL~\cite{29} model parameters. In this work, we design both our TFMs and SKFMs with 2 blocks, where the embedded feature dimension $E$ is 256, the number of full-body joints $J$ is 22, and the latent feature dimension for each joint $D$ is 64. To achieve better performance, we set the input sequence length $L$ to 96 frames. The network is trained on an NVIDIA 4090 device with a batch size of 256. For parameter optimization, we use the Adam solver~\cite{37}, and weight decay is set to 1e-5 during the whole training. The learning rate is set to 3e-4 and decays to 3e-5 after 200000 iterations. Our model achieves fast inference, taking only 12.9 ms to process 96 frames on an NVIDIA 4090 device.

\subsection{Evaluation Metrics} 
Following~\cite{6}, we assess model performance across three distinct metric categories. The first category measures rotational accuracy and is represented by the Mean Per Joint Rotation Error (MPJRE) [degrees]. The second category focuses on motion smoothness, which includes the Mean Per Joint Velocity Error (MPJVE) [cm/s] and Jitter ($10^{2}$$m/s^{3}$)~\cite{12}. The third category evaluates positional accuracy and includes the Mean Per Joint Position Error (MPJPE) [cm], Root PE, Hand PE, Upper PE, and Lower PE.

\begin{table}[t]
    \centering
    \caption{Evaluation results on AMASS under Protocol 2. }
    \label{tab:Table2}
    \small
    \begin{tabular}{l|cccc}
    \hline Method & MPJRE↓ & MPJPE↓ & MPJVE↓ & Jitter↓ \\
    \hline 
    % VAE-HMD$^{\dagger}$~\cite{17} & - & 7.45 & - & - \\
    % HUMOR$^{\dagger}$~\cite{38} & - & 5.50 & - & - \\
    % FLAG$^{\dagger}$~\cite{18} & - & 4.96 & - & - \\
    % \hline \hline
    AvatarPoser & 4.70 & 6.38 & 34.05 & 10.21 \\
    AvatarJLM & \underline{4.30} & \textbf{4.93} & 26.17 & 7.19 \\
    AGRoL & \underline{4.30} & 6.17 & \underline{24.40} & 8.32 \\
    SAGE & 4.62 & 5.86 & 33.54 & \underline{7.13} \\
    HMD-Poser* & 4.36 & 5.60 & 29.32 & 7.46 \\
    RPM* & 5.37 & 7.19 & 29.27 & \textbf{3.48} \\
    % \rowcolor[gray]{0.94}
    % \textbf{KineST (Ours)} & \underline{4.31} & \underline{5.27} & \textbf{24.23} & \underline{6.94} \\
    \textbf{KineST (Ours)} & \textbf{4.28} & \underline{5.17} & \textbf{24.08} & 7.36 \\
    \hline GT & 0 & 0 & 0 & 2.93 \\
    \hline
    \end{tabular}
\end{table}

\begin{table}[t]
    \centering
    \caption{Evaluation results on the real-captured data under Protocol 3.}
    \label{tab:real_data}
    % \scriptsize
    \small
    \begin{tabular}{l|cccc}
    \hline Method & MPJRE↓ & MPJPE↓ & MPJVE↓ & Jitter↓ \\
    \hline 
    AvatarPoser & 7.28 & 11.22 & 31.67 & 12.87 \\
    AvatarJLM & {7.01} & {9.72} & 27.59 & 13.10 \\
    % \rowcolor[gray]{0.94}
    \textbf{KineST (Ours)} & \textbf{6.91} & \textbf{9.68} & \textbf{25.16} & \textbf{9.49} \\
    \hline
    \end{tabular}
\end{table}

\subsection{Evaluation Results} 
We follow the recent common practice~\cite{69,70} of using AMASS dataset with two different protocols. \textbf{In the first protocol}, the subsets of the AMASS dataset, including CMU~\cite{33}, BMLrub~\cite{34} and HDM05~\cite{35}, are split into 90\% training data and 10\% testing data. \textbf{In the second protocol}, a larger benchmark from AMASS~\cite{10} is utilized, including 12 subsets for training and the HumanEva~\cite{49} and the Transition~\cite{10} subsets for testing. To further assess the real-world applicability of our method, we introduce \textbf{a third protocol}, where the model is evaluated on real headset-and-controller data collected by AvatarJLM~\cite{5} in an online setting, and the training setup remains consistent with Protocol 2.
%~\cite{33,34,35,40,41,42,43,44,45,47,48,47,10}
%~\cite{33,34,35,40,41,42,43,44,45,47,48,49,47,29}

\subsubsection{\textbf{Quantitative Evaluation}} 
As shown in Table~\ref{tab:Table1}, Table~\ref{tab:Table2}, and Table~\ref{tab:real_data}, our method achieves the best overall performance compared to existing approaches, demonstrating both high accuracy and robustness across key metrics. Specifically, in Table~\ref{tab:Table1}, KineST achieves a 2.59\% reduction in MPJRE, an 11.18\% decrease in MPJPE, and a 14.65\% improvement in MPJVE compared to MMD. Furthermore, as shown in Table~\ref{tab:Table2}, KineST achieves the lowest MPJRE and MPJVE, indicating its ability to construct precise and smooth motions in various actions. Although RPM achieves the best result in jitter, it exhibits significantly reduced pose accuracy. Notably, in Table~\ref{tab:Table2}, AvatarJLM achieves competitive MPJPE, and SAGE reports slightly lower jitter. However, they both rely on complex architectures with large parameter counts (63M and 137M respectively), which limits their practicality for lightweight AR/VR deployment. As shown in Table~\ref{tab:real_data}, our method outperforms all other methods across all metrics, demonstrating the practical applicability in real-world scenarios.

\subsubsection{\textbf{Qualitative Evaluation}} 
Visualization results are presented in Figure~\ref{fig.single_pose} and Figure~\ref{fig:seq_pose}. In Figure~\ref{fig.single_pose}, we compare the reconstruction errors of a single pose with other SOTA methods. In Figure~\ref{fig:seq_pose}, continuous motion sequence results are shown, with samples taken every 120 frames. The visualization results demonstrate that the proposed method achieves performance closest to the ground truth (GT).

\subsection{Ablation Study} 
In this section, the ablation studies are conducted on the AMASS dataset following the protocol 1 setting to verify the impact of each component and parameter in our designed model. The results are shown in Table~\ref{tab:Table4}-\ref{tab:Table6}.

% \begin{table}[t]
%     \centering
%     \caption{Evaluating the effect of different components in the Flow Module.}
%     \label{tab:Table3}
%     \scriptsize
%     \begin{tabular}{lcccccccc}
%     \hline
%     Method & MPJRE↓ & MPJPE↓ & MPJVE↓  & Jitter↓ \\
%     \hline
%     w/o GMA & 2.46 & 3.16  &  17.35  & 6.85  \\          
%     w/o LMA  & 2.27 & 2.90 & 16.99 & 7.75 \\   
%     w/o Bi-SSD block  & 2.37 & 3.04 & 20.70 & 13.57 \\
%     \textbf{Ours}  & \textbf{2.25} & \textbf{2.86} & \textbf{15.26} & \textbf{5.97} \\ 
%     \hline
%     \end{tabular}
% \end{table}

% Flow module结构消融
% \subsubsection{\textbf{Effects of different components in Flow Module}} We evaluate the significant role of each block in the proposed Flow Module, as shown in Table~\ref{tab:Table3}. The results reveal that the Bi-SSD block, LMA, and GMA each play a crucial and complementary role, and removing any of them leads to a clear performance drop across all metrics.
\begin{table}
    \centering
    \caption{Evaluating the effect of different scanning strategies.}
    \label{tab:Table4}
    \small
    \setlength{\tabcolsep}{4pt}
    \begin{tabular}{l|cccccccc}
    \hline
    Method & MPJRE↓ & MPJPE↓ & MPJVE↓  & Jitter↓ \\
    \hline
    Index-order in SMPL & 2.32 & 3.11  &  17.81  & 8.27  \\          
    FKS (Ours)  & 2.28 & 3.00 & 16.25 & 7.01 \\   
    % \rowcolor[gray]{0.94}
    \textbf{UKS (Ours)} & \textbf{2.25} & \textbf{2.86} & \textbf{15.26} & \textbf{5.97}\\
    \hline
    \end{tabular}
\end{table}

%  我们期望pure temporal更差
\begin{table}
    \centering
    \caption{Evaluating the effect of different modeling mechanisms in SKFM.}
    \label{tab:Table5}
    \small
    \setlength{\tabcolsep}{2.5pt}
    \begin{tabular}{l|cccccccc}
    \hline
    Method & MPJRE↓ & MPJPE↓ & MPJVE↓  & Jitter↓ \\
    \hline
    Pure Temporal & 2.27 & 2.97 &  16.84  & 7.83  \\          
    Pure Spatial (holistic)  & 2.41 & 3.10 & 16.77 & 7.72 \\   
    Pure Spatial (token-wise) & \textbf{2.23} & 2.93 & 17.85 & 9.31 \\  
    % \rowcolor[gray]{0.94}
    \textbf{STMM (Ours)} & 2.25 & \textbf{2.86} & \textbf{15.26} & \textbf{5.97}\\
    \hline
    \end{tabular}
\end{table}

\begin{table}
    \centering
    \caption{Evaluating the effect of the loss function.}
    \label{tab:Table6}
    \small
    \setlength{\tabcolsep}{4pt}
    \begin{tabular}{l|cccccccc}
    \hline
    Method & MPJRE↓ & MPJPE↓ & MPJVE↓  & Jitter↓ \\
    \hline
    Baseline & 2.25 & 2.87  &  16.10  & 6.75  \\  
    with $L_{angvel}^{diff}$ & 2.29 & 3.03  &  15.91  & 6.44  \\ 
    % \rowcolor[gray]{0.94}
   \textbf{with $L_{angvel}^{geo}$(Ours)} & \textbf{2.25} & \textbf{2.86} & \textbf{15.26} & \textbf{5.97}\\
    \hline
    \end{tabular}
\end{table}

% scan strategy ablation
\subsubsection{\textbf{Effects of different scanning strategies}} We compare three distinct strategies, including the index-order scan in SMPL, which sequentially scans from index 0 to 21, and the two proposed kinematics-guided bidirectional scans, FKS and UKS. As shown in Table~\ref{tab:Table4}, by embedding kinematic priors, FKS and UKS can effectively improve performance in joint relationship extraction. Additionally, due to the branch-wise design that harms the integrity of the human body, FKS cannot achieve better motion smoothness compared to UKS.

\subsubsection{\textbf{Effects of different modeling mechanisms in SKFM}} 
To evaluate the effectiveness of the proposed STMM in SKFM, we compare it with two other modeling mechanisms: pure temporal modeling and pure spatial modeling. Note that prior works have adopted two representative types of pure spatial modeling. One approach adopts a holistic modeling method that processes the concatenated features of all joints as a whole~\cite{71}, while the other follows a token-wise method that treats each of the 22 joints as an independent token to extract spatial features~\cite{5,69}. As shown in Table~\ref{tab:Table5}, the proposed STMM achieves a favorable balance between pose accuracy and motion smoothness compared to the other methods. While the token-wise spatial modeling yields the lowest MPJRE, independently extracting features for each joint inevitably overlooks the temporal continuity.

% loss ablation
\subsubsection{\textbf{Effects of the loss function}} We employ three distinct loss functions to train KineST. Since $L_{\text{rot}}$ and $L_{\text{ori}}$ are validated in previous works~\cite{3,5}, we take them as the baseline and focus on assessing the role of the proposed geometric angular velocity loss $L_{angvel}^{geo}$. Unlike calculating angular velocity using first-order finite differences $(L_{angvel}^{diff})$~\cite{70}, our proposed $L_{angvel}^{geo}$ operates it on the Lie group SO(3) in a geometrically consistent manner. As shown in Table~\ref{tab:Table6}, although $L_{angvel}^{diff}$ helps reduce MPJVE and Jitter, it significantly increases the errors in rotation and position. In contrast, the proposed $L_{angvel}^{geo}$ achieves smoother motion while preserving the performance on MPJRE and MPJPE, ensuring a good balance between accuracy and smoothness.

% loss没写
\section{Conclusion}
A novel kinematics-guided state space model, KineST, is proposed for full-body motion tracking from sparse signals, which mainly relies on two key innovations, the kinematics-guided bidirectional scanning strategy and the mixed spatiotemporal representation learning mechanism. To further improve motion continuity, a geometric angular velocity loss is designed to regulate rotational variations in manifold space. Extensive experiments show that KineST achieves superior accuracy and temporal consistency within a lightweight framework. We believe that the proposed model can significantly contribute to an immersive and seamless user experience in AR/VR applications.
\subsubsection{Limitation} A notable limitation of our approach, as well as related methods, lies in reconstructing relatively complex motions such as acrobatics or gymnastics. Future work will focus on incorporating richer motion priors during training, while maintaining low-cost and efficient inference.

\section{Acknowledgments}

This work was partly supported by the National Natural Science Foundation of China (Grant No.82472116) and the Natural Science Foundation of Shanghai (Grant No.24ZR1404100).

%We believe that the proposed model can significantly enhance the user experience in AR/VR applications.and seamless and dynamic
% \subsubsection{Limitation} A notable limitation of our approach, as well as related methods, lies in reconstructing relatively complex motions such as dancing or swimming. Future work will focus on incorporating rich motion priors during training, while maintaining low-cost and efficient inference.

% \subsection{References}

% \begin{quote}
% \begin{small}
% \textbackslash bibliography\{bibfile1,bibfile2,...\}
% \end{small}
% \end{quote}

% \appendix
% \section{Acknowledgments}

% % Anonymous submission version - shorter acknowledgments
% AAAI is especially grateful to Peter~\cite{1} Patel Schneider for his work in implementing the aaai2026.sty file, liberally using the ideas of other style hackers, including Barbara Beeton. We also acknowledge with thanks the work of George Ferguson for his guide to using the style and BibTeX files --- which has been incorporated into this document --- and Hans Guesgen, who provided several timely modifications, as well as the many others who have, from time to time, sent in suggestions on improvements to the AAAI style. We are especially grateful to Francisco Cruz, Marc Pujol-Gonzalez, and Mico Loretan for the improvements to the Bib\TeX{} and \LaTeX{} files made in 2020.

% Note: \bibliographystyle{aaai2026} is automatically set by aaai2026.sty
% Do not add \bibliographystyle{aaai2026} here as it will cause "Illegal, another \bibstyle command" error
\bibliography{aaai2026}
\clearpage
\appendix
% \section{Supplementary Material}
\begin{center}
    \LARGE \textbf{Supplementary Material}
\end{center}

% ----------- Supplementary Content Starts Here -----------

\section{Related Works}
\subsection{State Space Models }
State space models (SSMs)~\cite{20,25} recently gain prominence for sequential data processing. For instance, the structured state-space sequence model (S4)~\cite{20} is introduced to normalize parameters by a diagonal structure to analyze time series. Subsequently, the development of the S5 layer~\cite{25} and gated state space layer (GSS)~\cite{46} improves the efficiency of S4 by enhancing the parallel scanning and gated state space layer, respectively. More recently, Mamba~\cite{9} and Mamba-2~\cite{8} demonstrate competitive performance, faster inference speeds, and linear scaling capabilities with constant memory usage. As a result, various Mamba-based approaches emerge within computer vision~\cite{23,24}. In this work, the core framework of Mamba-2, named State Space Duality (SSD), is explored to estimate full-body motion from sparse observations. We enhance SSD with kinematic priors and mixed spatiotemporal representation learning for realistic full-body reconstruction.

\section{Preliminary}
\subsubsection{State Space Models} SSMs are a class of models for describing dynamic systems, widely used in signal processing and sequence modeling. Inspired by continuous linear time-invariant (LTI) systems, an input sequence $x(t) \in \mathbb{R}$ is transformed into $y(t) \in \mathbb{R}$ through an implicit latent state $h(t) \in \mathbb{R}^N$. The latent state $h(t)$ is associated with several matrix parameters $A\in \mathbb{R}^{N \times N}$, $B \in \mathbb{R}^{N \times 1}$, and $C \in \mathbb{R}^{N \times 1}$, and is governed by linear ordinary differential equations (ODEs):
\begin{equation}
h^{\prime}(t) = A h(t) + B x(t), \quad y(t) = C h(t).
\end{equation}
In practice, continuous-time SSMs are typically discretized via the zero-order hold (ZOH)~\cite{6} into a classic linear recurrence of the form:
\begin{equation}
h_t = \overline{A} h_{t-1} + \overline{B} x_t, \quad y_t = C h_t.
\end{equation}

% \begin{figure}
%     \centering
%     \includegraphics[width=1\linewidth]{LGMA2.png}
%     \caption{The architecture of the Local Motion Aggregator (LMA) and Global Motion Aggregator (GMA).}
%     \label{fig:lgma}
% \end{figure}

\section{Flow Module Details}
The Flow Module consists of a bidirectional SSD block (Bi-SSD), a Local Motion Aggregator (LMA), and a Global Motion Aggregator (GMA). Bi-SSD captures bidirectional temporal or spatiotemporal motion features, while LMA and GMA are designed to model short-range local dependencies and long-range periodic motion patterns, respectively.

\subsection{Local Motion Aggregator}
\begin{figure}[t]
    \centering
    \includegraphics[width=1\linewidth]{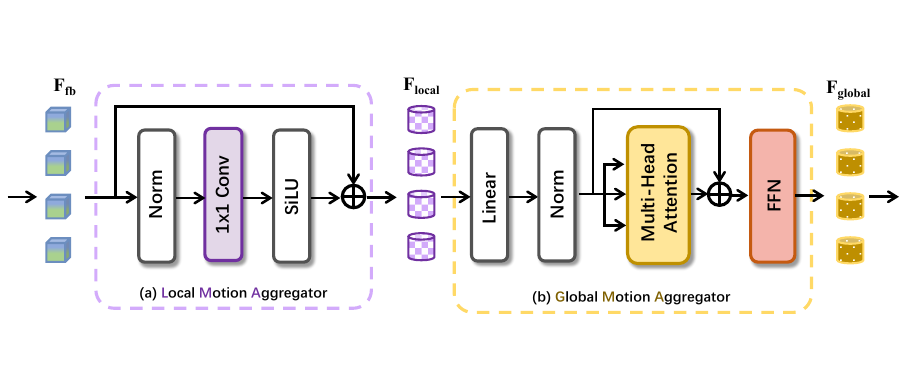}
    \caption{The architecture of the Local Motion Aggregator (LMA) and Global Motion Aggregator (GMA).}
    \label{fig:lgma}
\end{figure}

LMA is a convolution-based network, which takes $F_{fb} = F_f + F_b$ as input, where $F_f$ and $F_b$ denote the outputs of forward and backward SSD blocks, respectively. LMA consists of layer normalization, a 1D convolution, and a SiLU activation function, as shown in Figure~\ref{fig:lgma}(a). Specifically, the convolution layer is implemented as a 1D convolution with a kernel size of 1, which performs a per-frame linear transformation along the temporal dimension. This design allows the model to capture local temporal dependencies at each frame and outputs the local motion features $F_{local}$.

\subsection{Gocal Motion Aggregator}
GMA is a lightweight transformer-based network composed of a linear projection, layer normalization, multi-head self-attention, and a feed-forward network, as illustrated in Figure~\ref{fig:lgma}(b). Specifically, the local motion features $F_{local}$ are first linearly projected and normalized to ensure stable feature distribution. These features are then processed by a single-layer multi-head self-attention module~\cite{19}, with the hidden dimension constrained to 512 to reduce computational overhead. Finally, a feed-forward network (FFN) is applied to enhance feature representation to obtain the globally aggregated motion features $F_{global}$.

% \begin{figure*}
%     \centering
%     \includegraphics[width=1\linewidth]{pose.png}
%     \caption{Visualization results of different actions compared with other methods.}
%     \label{fig:placeholder}
% \end{figure*}

\section{Analysis of Former Angular Velocity Loss}
To alleviate temporal jitter in pose estimation, prior works have introduced angular velocity regularization by penalizing the discrepancy between predicted and ground-truth angular velocities. However, such approaches~\cite{70,72} typically estimate angular velocity by applying first-order finite differences on rotation representations. The formulation is defined as follows:

\begin{equation}
\begin{aligned}
V_{t} &= z_{t} - z_{t-1}, \\
\hat{V}_{t} &= y_{t} - y_{t-1}, \\
\mathcal{L}_{\text{angvel}}^{\text{diff}} &= \sum_{t=1}^{T-1} \left\| V_{t} - \hat{V}_{t} \right\|_{1}.
\end{aligned}
\end{equation}
Here, following the notation in the main text, $z_t$ denotes the rotation representation at frame $t$ computed from the ground-truth poses, $y_t$ represents the corresponding predicted rotation representation, $V_t$ denotes the angular velocity at frame $t$ computed from the ground-truth poses, and $\hat{V}_t$ represents the corresponding predicted angular velocity. 
Importantly, regardless of whether the rotation is parameterized using 6D vectors or axis-angle representations, directly computing differences in these representations fails to capture the true geometric relationship between rotations. Since rotations lie on the nonlinear manifold of the Lie group SO(3), angular velocity—defined as the time derivative of rotation—must be computed within the tangent space of SO(3), i.e., its Lie algebra $\mathfrak{so}(3)$. Differencing rotation features in Euclidean space and using them as angular velocity surrogates thus lacks a sound geometric foundation and does not reflect physically meaningful temporal dynamics.

\begin{table}[t]
    \centering
    \caption{Evaluating the effect of different components in the Flow Module.}
    \label{tab:Table3}
    \small
    \setlength{\tabcolsep}{4pt}
    \begin{tabular}{l|cccccccc}
    \hline
    Method & MPJRE↓ & MPJPE↓ & MPJVE↓  & Jitter↓ \\
    \hline
    w/o GMA & 2.46 & 3.16  &  17.35  & 6.85  \\          
    w/o LMA  & 2.27 & 2.90 & 16.99 & 7.75 \\   
    w/o Bi-SSD block  & 2.37 & 3.04 & 20.70 & 13.57 \\
    \textbf{Ours}  & \textbf{2.25} & \textbf{2.86} & \textbf{15.26} & \textbf{5.97} \\ 
    \hline
    \end{tabular}
\end{table}

\begin{table}[t]
    \centering
    \caption{Evaluating the effect of different loss functions.}
    \label{tab:loss}
    \small
    \setlength{\tabcolsep}{4pt}
    \begin{tabular}{l|cccccccc}
    \hline
    Method & MPJRE↓ & MPJPE↓ & MPJVE↓  & Jitter↓ \\
    \hline
    Base & 2.25 & 2.87  &  16.10  & 6.75  \\          
    with $\mathcal{L}_{\text{pos}}$  & 2.26 & 2.98 & 16.51 & 6.86 \\   
    with $\mathcal{L}_{\text{vel}}$  & 2.29 & 3.02 & 17.29 & 6.72 \\
    \textbf{with $L_{angvel}^{geo}$(Ours)}  & \textbf{2.25} & \textbf{2.86} & \textbf{15.26} & \textbf{5.97} \\ 
    \hline
    \end{tabular}
\end{table}

\begin{table}[t]
    \centering
    \caption{Evaluating the effect of different sequence lengths.}
    \label{tab:seqlen}
    \small
    \setlength{\tabcolsep}{4pt}
    \begin{tabular}{l|cccccccc}
    \hline
    Input Sequence Length & MPJRE↓ & MPJPE↓ & MPJVE↓  & Jitter↓ \\
    \hline
    {41} & 2.32 & 3.23  &  18.99  & 8.59  \\          
    \textbf{96(Ours)}  & \textbf{2.25} & \textbf{2.86} & \textbf{15.26} & \textbf{5.97} \\   
    {144}  & 2.47 & 3.42  &  18.78  & 7.47 \\
    {196}  & 2.40 & 3.21 & 17.46 & 6.51 \\ 
    \hline
    \end{tabular}
\end{table}

\section{Ablation Study}

\subsubsection{\textbf{Effects of different components in Flow Module}} We evaluate the significant role of each block in the proposed Flow Module, as shown in Table~\ref{tab:Table3}. The results reveal that the Bi-SSD block, LMA, and GMA each play a crucial and complementary role, and removing any of them leads to a clear performance drop across all metrics.

\subsubsection{\textbf{Effects of different losses}}
In addition to the proposed geometric angular velocity loss $\mathcal{L}_{\text{angvel}}^{\text{geo}}$, we also explore the positional loss $\mathcal{L}_{\text{pos}}$ and velocity loss $\mathcal{L}_{\text{vel}}$ during training, following prior works~\cite{3,5}:

\begin{equation}
\begin{aligned}
\mathcal{L}_{\text{pos}} &= \frac{1}{N} \sum_{i=1}^{N} \left\| FK(z_i) - FK(y_i) \right\|_2^2
\end{aligned}
\end{equation}

\begin{equation}
\begin{aligned}
\mathcal{L}_{\text{vel}} = \frac{1}{N{-}1} \sum_{i=1}^{N{-}1} \left\| 
\left( FK(z_{i+1}) - FK(z_i) \right) \right. \\
\phantom{=} \left. - \left( FK(y_{i+1}) - FK(y_i) \right) \right\|_2^2
\end{aligned}
\end{equation}
Here, following the notation in the main text, $Y = \left\{ y_i \right\}_{i=1}^L \in \mathbb{R}^{L \times V}$ denotes the predicted full-body motion sequence, and $Z = \left\{ z_i \right\}_{i=1}^L \in \mathbb{R}^{L \times V}$ denotes the corresponding ground-truth motion sequence. The function $FK(\cdot)$ denotes the forward kinematics function, which takes local joint rotations as input and outputs the global joint positions.

We evaluate different loss functions on our model, as shown in Table~\ref{tab:loss}. While $\mathcal{L}_{\text{angvel}}^{\text{geo}}$ and $\mathcal{L}_{\text{vel}}$ show effectiveness in previous literature, they do not lead to performance improvements in our framework.

We hypothesize that this discrepancy arises from the different modeling assumptions: prior methods do not explicitly incorporate human skeletal priors, making such geometric losses serve as useful compensatory constraints. In contrast, our model is designed with joint-chain structural priors embedded. Thus, the $\mathcal{L}_{\text{pos}}$ and $\mathcal{L}_{\text{vel}}$ may interfere with the learning of angular constraints, which is more significant.

\subsubsection{\textbf{Sequence length ablation}} Following the main setting, our method takes a sparse input signal 
$X=\left\{ x_i \right\}_{i=1}^L \in \mathbb{R}^{L \times C}$ and predicts a full-body pose sequence 
$Y=\left\{ y_i \right\}_{i=1}^L \in \mathbb{R}^{L \times V}$ of the same length $L$. 
In this experiment, we evaluate the model performance under four different sequence lengths. 
As shown in Table~\ref{tab:seqlen}, the model achieves the best performance across all three key metrics 
when the sequence length is set to 96.

% \section{Additional Visualizations}
% We present additional visualization evaluation compared with other SOTA methods, as illustrated in Figure~\ref{fig:placeholder}.

% ----------- Supplementary Content Ends Here -----------

% References and End of Paper
% These lines must be placed at the end of your paper
% \bibliography{aaai2026}

% \input{checklist/ReproducibilityChecklist}

\end{document}